\documentclass{ecai} 

\usepackage{microtype}
\usepackage{graphicx}
\graphicspath{{figures/}}
\usepackage[table]{xcolor}
\usepackage{booktabs} % for professional tables
\usepackage{colortbl}
\usepackage{subcaption}
\usepackage{multirow}

\usepackage{hyperref}
\usepackage{amsmath}
\usepackage{amssymb}
\usepackage{mathtools}
\usepackage{amsthm}

\usepackage{tikz}
\usetikzlibrary{positioning, shapes.geometric, arrows, fit, backgrounds, calc,patterns,angles,quotes}

\definecolor{group2}{RGB}{220,240,255} % Light Blue for first group
\definecolor{group2_alt}{RGB}{200,230,255} % Alternate Light Blue
\definecolor{group2_multicol}{RGB}{15,30,100} % Darker Blue for multicolumn

\definecolor{group1}{RGB}{255,230,200} % Light Orange for second group
\definecolor{group1_alt}{RGB}{255,220,180} % Alternate Light Orange
\definecolor{group1_multicol}{RGB}{100,30,10} % Darker Orange for multicolumn

\definecolor{group0}{RGB}{255,255,255} %white

\theoremstyle{plain}

\theoremstyle{definition}

\theoremstyle{remark}

\newcommand{\term}[1]{\emph{#1}}
\newcommand{\quotes}[1]{`#1'}

\begin{document}

\begin{frontmatter}
\title{seqKAN: Sequence processing with \\ Kolmogorov-Arnold Networks}
\author{\fnms{Tatiana}~\snm{Boura}}
\author{\fnms{Stasinos}~\snm{Konstantopoulos}}
\address{Institute of Informatics and Telecommunications,
  \\ NCSR~\quotes{Demokritos}, Ag.~Paraskevi, Greece
  \\ \texttt{\{tatianabou,konstant\}@iit.demokritos.gr}
}

\begin{abstract}

Kolmogorov-Arnold Networks (KANs) have been recently proposed as a
machine learning framework that is more interpretable and controllable
than the multi-layer perceptron. Various network architectures
have been proposed within the KAN framework targeting different tasks
and application domains, including sequence processing.
This paper proposes seqKAN, a new KAN architecture for sequence
processing.  Although multiple sequence processing KAN architectures
have already been proposed, we argue that seqKAN is more faithful
to the core concept of the KAN framework. Furthermore, we empirically
demonstrate that it achieves better results.
The empirical evaluation is performed on generated data from a complex
physics problem on an interpolation and an extrapolation task.
Using this dataset we compared seqKAN against a prior KAN network for
timeseries prediction, recurrent deep networks, and symbolic
regression. seqKAN substantially outperforms all architectures,
particularly on the extrapolation dataset, while also being the most transparent.


\end{abstract}

\end{frontmatter}

\section{Introduction}
\label{sec:intro}

\term{Kolmogorov-Arnold Networks (KANs)} were recently proposed by
\citet{liu-wang-etal:2024} as an alternative machine learning
framework to the ubiquitous \term{multi-layer perceptron (MLP)}.
KANs introduce the idea that if edge weights are lifted to
learnable functions, this suffices to capture non-linearities in
the data so that nodes can simply sum incoming edges.
This idea is inspired by the
\term{Kolmogorov-Arnold Representation Theorem (KAT)} that proves that
any multi-variate function can be re-formulated using two layers of
uni-variate functions and simple (unweighted) summation.

KAT has been the object of an extensive discussion regarding its
relevance to machine learning, summarized by \citet{schmidt:2021}.
One notable point in this discussion is that KAT guarantees the
existence of uni-variate functions that can be combined into an
exact representation any multi-variate function, but offers neither a
construction nor any guarantee on the properties of these uni-variate
functions other than being continuous. In fact, with the exception of
trivial cases, these functions are suspected to be highly non-smooth
and practically impossible to construct either analytically or
empirically.

The KAN architecture circumvents objections to KAT's relevance to
machine learning by framing KANs as \emph{approximators} (as opposed to
KAT's exact representations) that stack two \emph{or more} layers.
This re-contextualization into the modern deep learning environment
has mustered impressive interest from the machine learning community
with more than 10 extensions and applications published within the few
months since the original KAN article \citep[Section~1]{basina-etal:2024}.

In this paper we present \term{seqKAN}, a new architecture within the
KAN framework for processing sequences. Our main contribution is that
this architecture is more faithful to the core concept of the KAN
framework as it avoids re-introducing weighted summation and fixed
activation function in the form of conventional MLP-styled recurrency
cells. A secondary contribution is the definition of a new evaluation
task that has several characteristics (lacking from existing tasks)
geared towards combining quantitative evaluation with a qualititative
analysis of the results. In the remainder of
this article we will first provide the relevant background on KANs and
their recurrent extensions (Section~\ref{sec:bg}) and then present
seqKAN (Section~\ref{sec:seqkan}). We will then present and discuss
the evaluation task and our experimental results (Section~\ref{sec:exp})
and finally close with conclusions and directions for future work
(Section~\ref{sec:conc}).

\section{Background}
\label{sec:bg}

\subsection{Kolmogorov-Arnold networks}
\label{sec:kan}

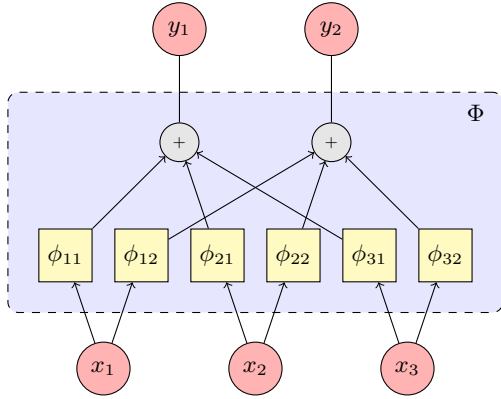
\begin{figure}[t]
\begin{center}

\begin{tikzpicture}[
    node distance=1cm and 1cm,
    input/.style={circle, draw, fill=red!30, minimum size=0.7cm},
    middle/.style={rectangle, draw, fill=yellow!30, minimum size=0.7cm},
    sum/.style={circle, draw, fill=gray!20, minimum size=0.2cm, inner sep=0.1cm},
    output/.style={circle, draw, fill=red!30, minimum size=0.7cm},
    every node/.style={font=\small}
]

\node[input] (i1) at (0, 0.5) {$x_1$};
\node[input] (i2) at (2, 0.5) {$x_2$};
\node[input] (i3) at (4, 0.5) {$x_3$};

\node[middle] (m1) at (-0.5, 2) {$\phi_{11}$};
\node[middle] (m2) at (0.5, 2) {$\phi_{12}$};
\node[middle] (m3) at (1.5, 2) {$\phi_{21}$};
\node[middle] (m4) at (2.5, 2) {$\phi_{22}$};
\node[middle] (m5) at (3.5, 2) {$\phi_{31}$};
\node[middle] (m6) at (4.5, 2) {$\phi_{32}$};

\node[sum] (s1) at (1, 3.5) {\tiny{$+$}};
\node[sum] (s2) at (3, 3.5) {\tiny{$+$}};

\node[output] (o1) at (1, 5.) {$y_1$};
\node[output] (o2) at (3, 5.) {$y_2$};

\begin{scope}[on background layer]
    \node[draw, dashed, rounded corners, fill=blue!10, fit=(m1) (m2) (m3) (m4) (m5) (m6) (s1) (s2), inner sep=0.4cm, label={[xshift=1.2cm,yshift=-0.5cm]above right:$\Phi$}] {};
\end{scope}

\draw[->] (i1) -- (m1);
\draw[->] (i1) -- (m2);

\draw[->] (i2) -- (m3);
\draw[->] (i2) -- (m4);

\draw[->] (i3) -- (m5);
\draw[->] (i3) -- (m6);

\draw[->] (m1) -- (s1);
\draw[->] (m2) -- (s2);
\draw[->] (m3) -- (s1);
\draw[->] (m4) -- (s2);
\draw[->] (m5) -- (s1);
\draw[->] (m6) -- (s2);

\draw[-] (s1) -- (o1);
\draw[-] (s2) -- (o2);

\end{tikzpicture}

\caption{An example of a 2-layer KAN with three input nodes fully connected
  to two output nodes. The value at each input node is passed to two
  activation functions, one for each output node. For instance, for
  input node $x_1$, there are two $\phi_{1x}$ activation functions:
  $\phi_{11}$ contributes to output node $y_1$ and $\phi_{12}$
  contributes to output node $y_2$.
  The value at each output node is the sum of the results of the
  corresponding activation functions. So the value of output node $y_1$
  is $\phi_{11}(x_1)+\phi_{21}(x_2)+\phi_{31}(x_3)$
  and the value of $y_2$ is $\phi_{12}(x_1)+\phi_{22}(x_2)+\phi_{32}(x_3)$.}
\label{fig:kan_layer}
\end{center}
\vspace{1.5em}
\end{figure}

As mentioned above, the basic idea underlying KANs is that the
activation functions that transform values as they move forward
through layers are learned from the data and not pre-determined; and
that each node in a layer performs simple addition (without weights)
over the values it receives from each node of the preceding
layer. Figure~\ref{fig:kan_layer} gives a characteristic example of
how values from a 3-node layer propagate to a 2-node layer.

Each \term{activation function} $\phi_i(\cdot)$ is the weighted sum of
a learned spline and $\mathrm{silu}(x) = x/(1+\exp(-x))$. That is:
$$
\phi_i(x) = w_1\cdot\mathrm{silu}(x) + w_2\cdot\mathrm{spline}_i(x)
$$
where $w_1,w_2$ and $\mathrm{spline}_i$ need to be trained for each
edge of the network. The original publication explains $w_1,w_2$ as
implementation details that make the network well-optimizable
\citep[p.~6]{liu-wang-etal:2024}.
Besides completely learned splines, activation functions can also
be parameterized prior functions. In this case, the user provides
univariate function implementations which the network parameterizes
via affine transformations (translation and scaling) to fit the data.

By comparison to a conventional neural network, KANs have
significantly more parameters to train. Assuming the same number of
nodes and edges, a NN learns a single weight for each incoming edge of
each node. A KAN, by comparison, learns $w_1,w_2$ and the spline
parameters or the translation/scaling parameters for each edge.
As a consequence, KANs are a viable alternative only when they can
adequately approximate the target function with considerably fewer and
narrower layers, resulting in significantly fewer edges. When seen
under this light, KANs are a significantly \emph{less distributed}
representation than NNs. This is further emphasised by the fact that
the loss function biases training towards \term{sparsity}, that is,
towards having as few nodes as possible receive non-zero inputs and
produce non-zero outputs. Naturally, nodes with almost-zero inputs and
outputs are pruned to improve readability with minimal impact on
network performance. This technique is similar to regularization in
MLPs, but in KANs the result is directly interpretable.

The above make KANs both better \term{interpretable} and more directly
\term{controllable}. Interpretability basically boils down to the fact
that multiple parameters are lumped together into a spline, which can
be easily visualized as a function graph and understood by the
operator as a familiar function. This greatly reduces the number of
different objects and interactions between them that need to be
absorbed by a human operator in order to understand what has been learned.
Controllability in KANs has multiple facets: from the ability to provide
prior functions besides the learned ones, to the ability to replace a
familiar-looking spline with a known function, to the ability to prune
dependencies that appear to be overfitting or, in general, not
capturing the underlying phenomenon.

\subsection{Processing sequence data}

There are two major approaches in processing sequences: Those
receiving the complete sequence as input and those receiving the
sequence one datapoint at a time.
In the former case the system uses the position in the input
vector or, in some cases, an explicit time representation to encode
temporal relationships. The \term{Transformer} architecture
\citep{vaswani-etal:2017} is a prime example of this approach, where
\term{attention} is trained to weigh past tokens that are pertinent to
the processing of the current token.

Investigating how the concepts of attention and KAN can be integrated
can potentially give fruitful results, but the current KAN literature only
includes extensions of \term{Recurrent Neural Networks (RNN)} and,
in fact, of LSTM networks.
The \term{Long Short-Term Memory (LSTM)} \citep{hochreiter-schmidhuber:1997}
and the \term{Gated Recurrent Unit (GRU)} \citep{cho-etal:2014}
architectures are examples of the more specific family of
architectures within the RNN framework where \term{gates} are trained
to control the flow of information to and from a \term{hidden state}
that distills the effect of past inputs on the processing of the
current input.

The most mature among these initial approaches to utilizing KANs for
sequence processing is the
\term{Temporal Kolmogorov-Arnold Network (TKAN)} \citep{tkan:2024}.
TKAN replaces the output layer of the LSTM cell with an array of KANs,
each of which KANs itself comprises multiple KAN layers. The outputs
of these KANs are combined into a single vector through trainable
weights. This combined vector includes (a) a recurrent input that is
fed back into both the LSTM cell and the KAN layers, and (b) the
output of the TKAN layer. These layers are then stacked to form a
TKAN network.

\section{seqKAN}
\label{sec:seqkan}

Our seqKAN architecture introduces recurrency directly into the KAN
architecture without adding structures that rely on trained weights
and fixed activation functions. This makes seqKAN more faithful to the
core concept of the KAN framework, and more capable of fully
exploiting the interpretability and controllability offered by KANs.

By contrast, the way TKAN integrates KAN layers within the LSTM cell
\emph{and} also uses trained weights to combine values leaves a good
part of the overall TKAN architecture on the side of MLP networks and
outside the scope of learned activation functions. To a large extent
this undermines the value of using a KAN-based architecture:
We have no reason to believe that significant pieces of the knowledge
distilled from the data will not be stored in the MLP part of the
network and become opaque.
Furthermore, even the knowledge that is stored in the learned functions
will be obfuscated by its interaction with the opaque part of the
network offering no opportunities to understand or, even more so, edit
and control what has been learned.

This goes beyond TKAN and LSTM and is a more general statement of our
core motivation: The choice of activation functions in RNNs is the
result of careful consideration and extensive experimentation and has
proven itself effective. We have no reason to believe that replacing
these activation functions with learnable functions
\emph{in the same or a similar network structure} can improve results.
If anything, such a move introduces the overhead of fitting splines
instead of using the functions that are known to work well.

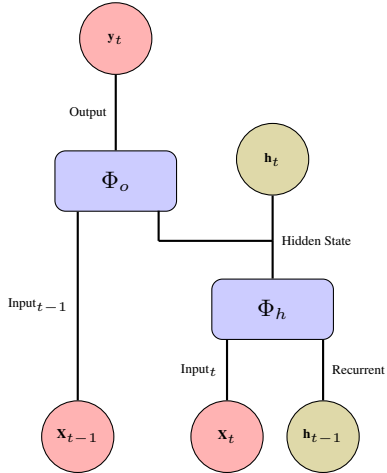
\begin{figure}[t]
\begin{center}
\begin{tikzpicture}[
    every node/.style={font=\small},
    block/.style={rectangle, draw, fill=blue!20, rounded corners, minimum width=1.6cm, minimum height=0.8cm},
    circleblock/.style={circle, draw, fill=black, minimum size=0.01cm},
    input/.style={coordinate},
    output/.style={coordinate},
    arrow/.style={->, thick},
    line/.style={-, thick},
    kaninput/.style={circle, draw, fill=red!30, minimum size=0.95cm},
    hidden/.style={circle, draw, fill=olive!30, minimum size=0.95cm},
    kanoutput/.style={circle, draw, fill=red!30, minimum size=0.95cm},
]

% Input, hidden state, and output
\node[kaninput] (xt) {\tiny{$\textbf{X}_t$}};
\node[block, above=0.8cm of xt] (phi1) [xshift=0.6cm, yshift=0.0cm] {$\Phi_{h}$};
\node[hidden, right=0.3cm of xt] (hprev) {\tiny{$\textbf{h}_{t-1}$}};
\node[kaninput, left=1.0cm of xt] (xt0) {\tiny{$\textbf{X}_{t-1}$}};
\node[output, above=1.0cm of phi1] (ht) {};
\node[output, above=0.8cm of xt] (ghostph1) {};

\node[output, above=0.8cm of hprev] (ghostph2) {};

\node[output, above=2.5cm of xt0] (ghostph3) {};
\node[block, above=2.5cm of xt0] (phi2) [xshift=0.5cm, yshift=0.0cm] {$\Phi_o$};

\node[kanoutput, above=1cm of phi2] (yt) {\tiny{$\textbf{y}_t$}};

\node[hidden, above=0.1cm of ht] (htta) {\tiny{$\textbf{h}_{t}$}};
\node[output, below=0.5cm of ht] (htt) {};
\node[output, left=1.5cm of htt] (httr) {};
\node[output, above=0.4cm of httr] (ghostph4) {};

% Connections
\draw[line] (xt) -- (ghostph1) node[midway, left] {\tiny{Input$_{t}$}};
\draw[line] (xt0) -- (ghostph3) node[midway, left] {\tiny{Input$_{t-1}$}};
\draw[line] (hprev) -- (ghostph2) node[midway, right] {\tiny{Recurrent}};
\draw[line] (htt) -- (httr) node[midway, above] {};
\draw[line] (htt) -- (htta) node[midway, above] {};
\draw[line] (httr) -- (ghostph4) node[midway, above] {};
\draw[line] (phi1) -- (htt) node[above, right] {\tiny{Hidden State}};
\draw[line] (phi2) -- (yt) node[midway, left] {\tiny{Output}};

\end{tikzpicture}

\caption{seqKAN architecture. At each timestep, the multivariate input \textbf{$X_t$}, along with the previous hidden
state \textbf{$h_{t-1}$}, is passed through the hidden-state layer \textbf{$\Phi_h$}, which computes the current hidden
state \textbf{$h_t$}. The current hidden state is then passed through the output layer \textbf{$\Phi_o$}, which computes
the current output by also utilizing the multivariate input from the previous timestep, \textbf{$X_{t-1}$}.
}
\label{fig:seqkan}
\end{center}
\vspace{1.5em}
\end{figure}

In order to force the network to store knowledge in the less
distributed, more interpretable KAN splines we should develop a KAN
architecture that is performant without incorporating MLP structures.
Our seqKAN architecture draws inspiration from the RNN architecture
but reformulates it in a purely KAN network without trainable weights.
seqKAN has one layer that receives input and maps it to a hidden state
that is both pushed back to the input and sent to an output layer that
computes the outputs (Figure~\ref{fig:seqkan}).

In the specific configuration used in our experiments, the output
layer is a [3,2] KAN layer, identical to the one shown in
Figure~\ref{fig:kan_layer}. The output layer $\Phi_o$ maps a
single-variable hidden state and the two variables from the previous
time-step to the two output variables. The hidden-state layer $\Phi_h$
is a [3,1] KAN layer that maps the two input variables and the
previous hidden state to the new hidden state.

The architectural decision to feed the previous inputs into the output
layer is driven by preliminary experiments that
compared this architecture against a seqKAN/wide architecture
where a wider $\Phi_h$ gave the network enough degrees of freedom to
learn identity activation functions in $\Phi_h$ to simply maintain the
previous inputs and pass them to the output layer. In these
preliminary experiments seqKAN outperformed seqKAN/wide by a large
margin. We will revisit this point in Section~\ref{sec:exp:seqkan}.

\section{Experimental Results and Discussion}
\label{sec:exp}

\subsection{Experimental Setup}
\label{sec:exp:exp_setup}

In order to evaluate seqKAN's ability to capture temporal dependencies
in multivariate time series inputs, we designed a new learning task
that has the following characteristics that we have not (cumulatively)
found in tasks previously used in the KAN or sequence processing
literature:
\begin{itemize}
\item It is a multi-variate sequence where the variables interact and
  affect each other (and, of course, the expected outputs)
  non-linearly, since the ability to disentangle and elucidate complex
  dependencies between the input variables is a key promise
  of KANs.
\item It models a non-stationary, evolving phenomenon where a
  superficial drift can be explained by discovering a long-term
  trend. Naturally, the interaction between the trend and the outputs
  should not be simple superposition but a non-linear
  composition. This tests both the ability to disentangle complex
  compositions as well as the ability to build models with radically
  different behaviours in different parts of the variables' value
  domains without catastrophic forgetting. This latter aspect, in
  particular, is meant to test the value KAN extract from fitting
  splines as opposed to using pre-defined functions.
\item It models a well-understood phenomenon and offers itself to an
  analysis that qualitatively evaluates if the learned network
  actually models the underlying phenomenon or has chanced upon a
  solution that fits the available data but is not expected to perform
  well in corner cases; Noting, of course, that well-understood only
  implies that the correct computation is known and not that it is
  straightforward.
\end{itemize}
Based on the above, we built our task on data from the dynamics of a
pendulum where the string gets longer as time progresses, resulting in
motion captured by differential equations without closed-form
solutions \citep{yakubu-etal:2022}.

More details on the physics of the problem and the exact equations
used to generate motion data are given in
Appendix~\ref{sec:pendulum}. For our purposes here, it suffices to
clarify that string length $L$ depends only on time. But $L$, the
angular displacement $\theta$ of the pendulum, and its derivatives
angular velocity $\omega$ and angular acceleration $\alpha$ form a
system of differential equations.
This system has no closed-form solution but can be solved numerically
by stepping through time from its initial parameters, which makes it
appropriate as a sequence processing task since future values can be
predicted from previous ones. Specifically, since the equations
involve the first two derivatives of displacement, the calculation
needs the second order of differences. Therefore from three time steps
and $L$ one can calculate the next time step.

However, there is the complication that as the sequence progresses its
dynamics change because $L$ changes. By withholding $L$ from the
inputs, we create a non-stationary sequence where failure to model the
long-term trend will result in considerably higher loss when testing on
data extracted from areas of the value space of $L$ not seen during
training. In other words, this setup gives the ability to generate an
\emph{interpolation} test set where unseen test data can be approximated
without capturing the real underlying dynamics and an \emph{extrapolation}
test set where such an approximation will result in high loss.
We use the exponential function
$$
L(t) = 0.1 \cdot \exp_{10}(\frac{5.88\cdot t}{1000})
$$
to calculate the length of the pendulum string.\footnote{In preliminary
  experiments we compared fixed, linear, and the (actually used)
  exponential length functions. As expected, the fixed-length task is
  too easy to make meaningful comparisons. In the linear-length task
  some differences between systems emerge, but are difficult to
  discern.}

From each timestep we extract two binary output variables from the
motion parameters: (a) whether the displacement is close to the
equilibrium position and (b) whether the energy is increasing or
decreasing.
The exact definitions are based on lecture notes by \citet{tedrake:2023}
and also given in Appendix~\ref{sec:pendulum}.
Here it suffices to state that the \emph{Close to Equilibrium} label
only needs to test whether $\theta$ and $\omega$ are same-sign or
opposite-sign and is straightforward to compute from the sequence whereas
the \emph{Energy Increasing} label is a more complex
computation also involving $L$.

Since the two labels share parts of the calculation we expect that
combining these labels in a joint learning task helps networks to
extract the essential underlying properties of the phenomenon instead
of overfitting on a single label. For this reason we defined loss as
the mean of the Binary Cross Entropy between the ground-truth value
and the value calculated by the network for each of the two labels.

We generated a training set of 200 datapoints, where each datapoint is
a sequence of ten $\left<\theta,\omega\right>$ pairs and the two labels
extracted from the motion parameters of the final pair. Similarly, we
generated an interpolation test set of 200 datapoints from the same
timesteps but with a different initial displacement.
Finally, we generated an
extrapolation test set of 200 datapoints from timesteps subsequent to
the last training datapoint. We chose the sequence length of ten to be
considerably longer than the required three, since the systems also
need to bring their hidden states to a point where the hidden
variable $L$ is estimated from the observed displacement and velocity
variables.

To compare the performance of our proposed model, we evaluate it against several sequential data processing models on
both interpolation and extrapolation tasks.
For baseline performance, we run experiments using an RNN and an LSTM.
To compare with a method that explicitly models functions, we employ symbolic regression.
Finally, we compare the seqKAN's performance with the TKAN model
to assess different KAN-based architectures.
All models are configured to use the same number of parameters.

For our experiments we used the following implementations:
\begin{itemize}
\item Our \mbox{PyTorch} implementation of seqKAN,
  \url{https://github.com/data-eng/seqKAN}
\item The reference implementation of TKAN at
  \url{https://github.com/remigenet/TKAN}
\item Our \mbox{PyTorch} implementations of RNN and LSTM.
\item The PySR~1.3.0 symbolic regressor at
  \url{https://github.com/MilesCranmer/PySR}
\end{itemize}

Our implementations of seqKAN, RNN, and LSTM and of the dataset
generator as well as the actual dataset used in the experiments
described here can be retrived from
\url{https://doi.org/10.5281/zenodo.14899652}

\subsection{Results with seqKAN}
\label{sec:exp:seqkan}

As previously mentioned in Section~\ref{sec:seqkan}, we tested two
variations of the seqKAN architecture in order to establish which one
performs better. The seqKAN architecture shown in
Figure~\ref{fig:seqkan} hard-wires the previous input variables as
inputs to the output layer, re-enforcing the effect of the
immediately previous input beyond what can be retained in the
single-variable hidden-state layer. This optimizes network size at the
expense of generality, since we directly allocated learnable functions
to treat the immediately previous inputs which we know to be of great
relevance.

The more general alternative which we shall call \mbox{\em seqKAN/wide}
in the results presented here, is to remove these direct inputs and
instead have a wider hidden-state layer. Such a wider layer has the
capacity to let two of its functions simply be $y=x$ to allow the
previous input to be retained verbatim in the hidden state.

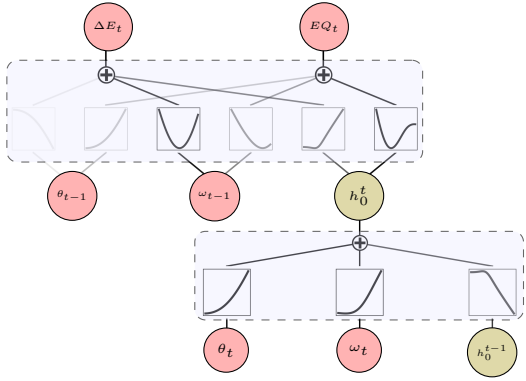
\begin{figure}[t]
\begin{center}
\begin{tikzpicture}[
    every node/.style={font=\tiny},
    block/.style={rectangle, draw, fill=blue!20, rounded corners, minimum width=1.6cm, minimum height=0.8cm},
    circleblock/.style={circle, draw, fill=black, minimum size=0.01cm},
    input/.style={coordinate},
    output/.style={coordinate},
    arrow/.style={->, thick},
    line/.style={-, thick},
    kaninput/.style={circle, draw, fill=red!30, minimum size=0.1cm},
    hidden/.style={circle, draw, fill=olive!30, minimum size=0.1cm},
    kanoutput/.style={circle, draw, fill=red!30, minimum size=0.1cm},
]

\node[] (phi1) {\includegraphics[width=5.5cm]{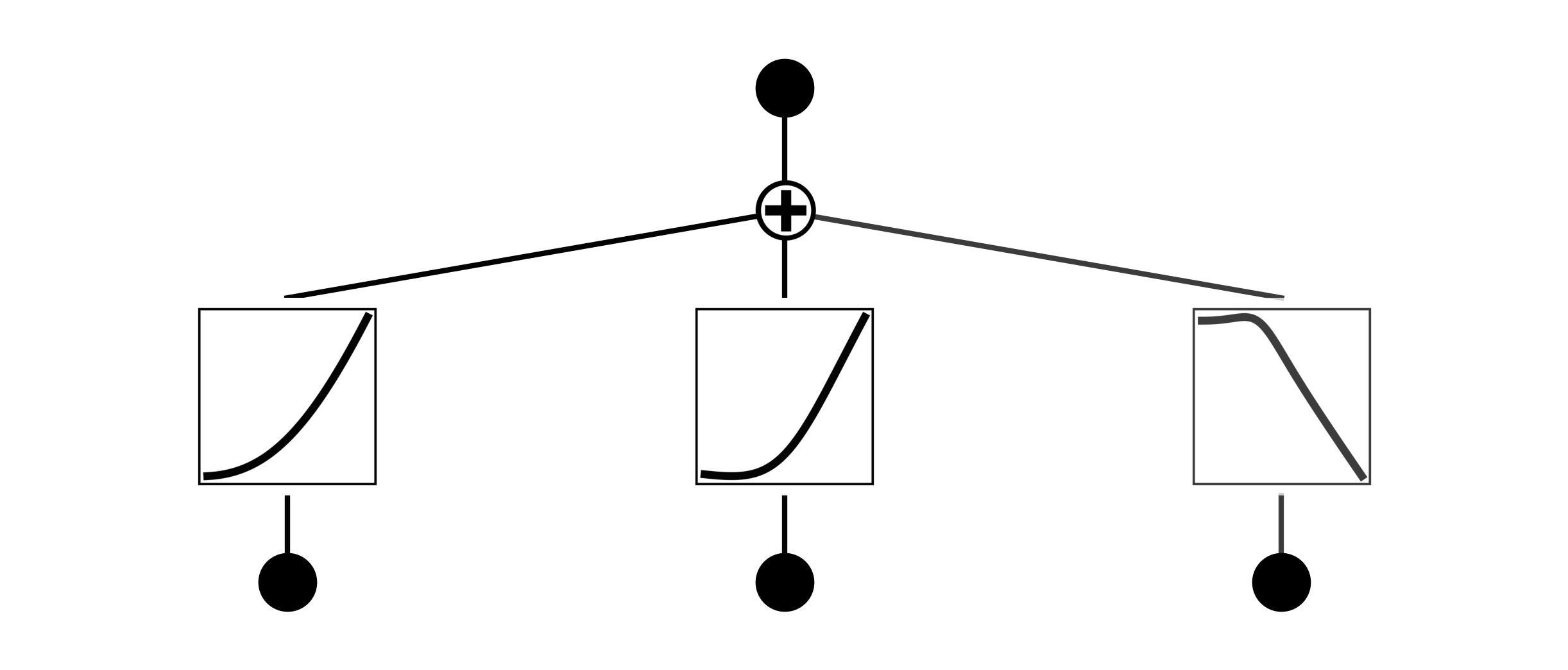}};

\node[above=-0.5cm of phi1] (phi2) [xshift=-1.9cm, yshift=0.0cm] {\includegraphics[width=6cm]{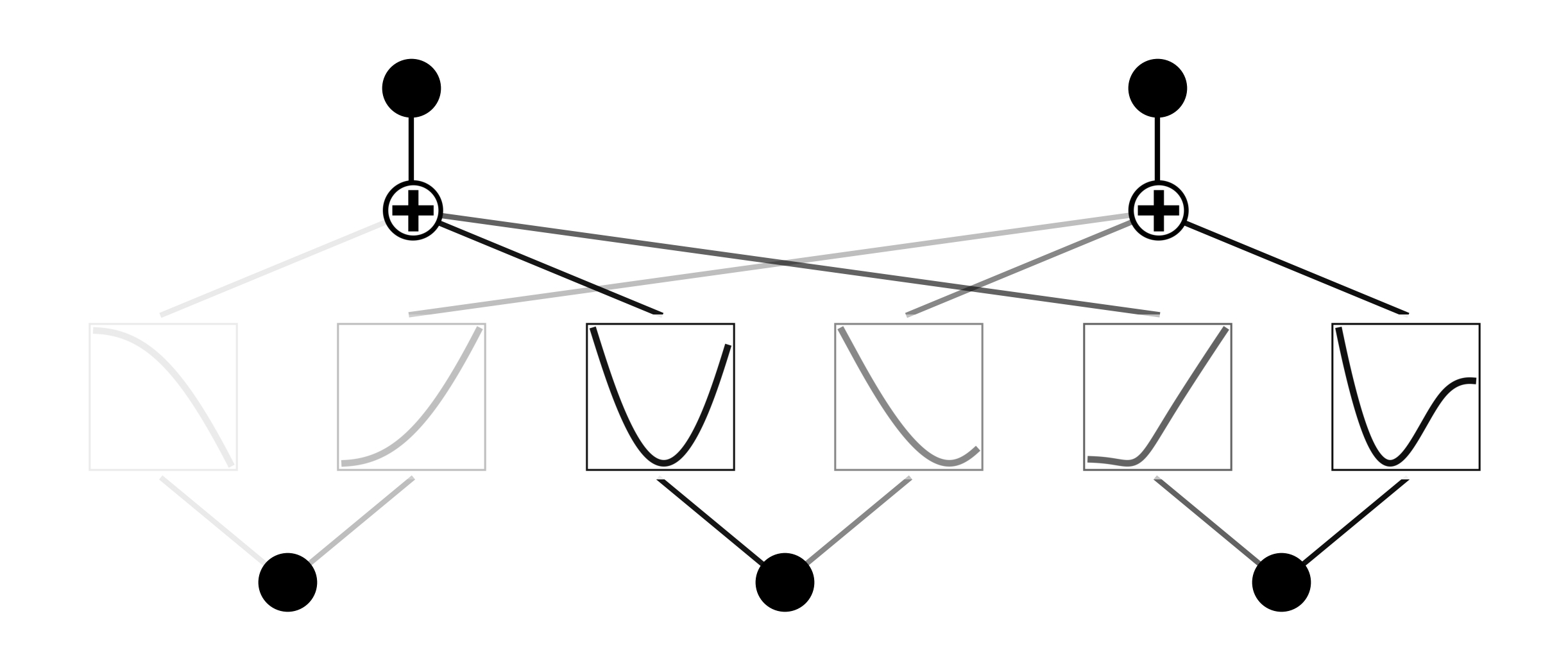}};

\node[kaninput, below=-0.6cm of phi1] (x0t) [xshift=-1.75cm, yshift=0.0cm] {$\theta_t$};
\node[kaninput, below=-0.6cm of phi1] (x1t) [xshift=0.0cm, yshift=0.0cm] {$\omega_t$};
\node[hidden, below=-0.6cm of phi1, scale=0.7] (h0t1) [xshift=2.5cm, yshift=0.0cm] {$h_0^{t-1}$};

\node[kaninput, below=-0.6cm of phi2, scale=0.75] (x0t1) [xshift=-2.5cm, yshift=0.0cm] {$\theta_{t-1}$};
\node[kaninput, below=-0.6cm of phi2, scale=0.75] (x1t1) [xshift=-0.0cm, yshift=0.0cm] {$\omega_{t-1}$};
\node[hidden, below=-0.6cm of phi2, scale=0.95] (h0t) [xshift=2.0cm, yshift=0.0cm] {$h_0^{t}$};

\node[kaninput, above=-0.6cm of phi2, scale=0.8] (y0t) [xshift=-1.75cm, yshift=0.0cm] {$\Delta E_t$};
\node[kaninput, above=-0.6cm of phi2, scale=0.8] (y1t) [xshift=1.8cm, yshift=0.0cm] {$EQ_t$};

\node[draw, dashed, rounded corners, fill=blue!10, fit=(phi1), inner sep=-0.7cm, fill opacity=0.3, draw opacity=0.6] {};
\node[draw, dashed, rounded corners, fill=blue!10, fit=(phi2), xscale=0.85, yscale=0.45, fill opacity=0.3, draw opacity=0.6] {};
\end{tikzpicture}

\caption{Trained seqKAN model. It takes current and previous $\omega$ and $\theta$ as inputs, has a single hidden
state $h_0$, and predicts  the labels for energy change ($\Delta E$) and equilibrium ($EQ$). The corresponding $\phi$
    functions are shown with varying opacities, where darker shades indicate a stronger influence on the prediction,
    while lighter shades signify a diminishing effect.}
\label{fig:seqkan_best_trained}
\end{center}
\vspace{1.5em}
\end{figure}

\begin{figure}[t]
\begin{center}
\includegraphics[width=\linewidth]{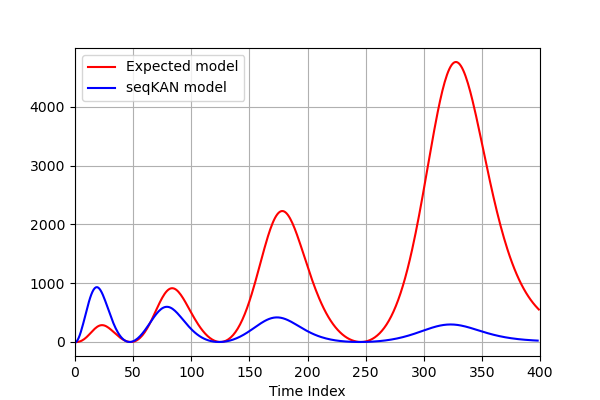}
\caption{Comparison of the priorly expected model and the model actually
  learned by seqKAN for the Energy-Increasing label.
  The seqKAN model is scaled by 10 for plotting clarity.}
\label{fig:seqkan_vs_reality}
\end{center}
\vspace{1.5em}
\end{figure}

The quantitative results in
Table~\ref{tab:combined_inter_extrapolation} show that
\mbox{seqKAN/wide} achieves a significantly lower performance,
especially in the most challenging task of predicting the Energy
Increasing label in extrapolated datapoints. But the great advantage
that the KAN framework is the ability to analyse not only the
quantitative results, but also the actual calculations that led to
these results. This will give a better understanding of what is
observed experimentally.

In our specific case, we can read in
Figure~\ref{fig:seqkan_best_trained} that the actual calculation for
the Energy-Increase label is an expression along the lines of
$\omega_{t-1}^2+(\Delta\theta_t)^2+(\Delta\omega_t)^3$, ignoring any
scaling that might be encoded in the actual splines and is not visible
in the figure. This was derived as follows:
\begin{itemize}
\item The $\Delta E_t$ node is the sum of a function similar
  to $\omega_{t-1}^2$ and a function similar to $h_o^t$ for larger
  time indexes, while the latter term is zeroed out for smaller time
  indexes.
\item The $-\theta_{t-1}$ term is considered inconsequential by the
  network can be safely ignored. The low significance score is
  indicated by the very light shade in the figure. At this point it is
  important to remind the reader that these scores are \emph{not}
  weights but are used to decide if a term will be included in the sum
  or completely removed to achieve network sparsity.
\item $h_o$ is the sum of three terms: $\theta_t$,
  $\omega_t$ and the opposite of the previous $h_o$, which in turn
  adds $\theta_{t-1}$, $\omega_{t-1}$, and the $h_o$ before that.
  By simply re-arranging the terms, this can be written as
  $\Delta\theta_t^2+\Delta\omega_t^3$
  and some residue from previous
  iterations. Noting that the function applied to $h_o^{t-1}$ appears
  linear but the functions applied to $\theta_t$ and $\omega_t$ appear
  polynomial we infer that the network aims at a residue that
  diminishes the effect of more temporally distant datapoints.
  So the expession
  $\Delta\left(\theta_t+c_1\right)^2 + \Delta\left(\omega_t+c_2\right)^3$
  can represent this term.\footnote{The constants $c_1$ and $c_2$ achieve
    the horizontal translation apparent in the plot.}
\end{itemize}
As can be seen in Figure~\ref{fig:seqkan_vs_reality},
there is almost perfect correlation between
$\omega_{t-1}^2+(\Delta\theta_t)^2+(\Delta\omega_t)^3$
and the expression
$\left(\sin(\theta_t)\cdot\omega_t/\Delta\omega_t\right)^2$,
which is a good approximation of the analytically-derived formula for
this label (cf. Appendix~\ref{sec:pendulum}).\footnote{Since the
  plots correlate, it is only a matter of getting scale and
  translation right to get the correct zero-crossings. Naturally, the
  exact splines can be retrieved after training, but that is not
  important for the qualitative argument being made here: We know that
  the scaling/translation in the exact splines is correct because of
  the quantitative results.}
It should be noted that the correlation holds for all the timesteps,
including the timesteps beyond 200 where the model extrapolates.
In other words, the model has correctly extracted the underlying
drift towards ever-increasing pendulum periods as the string gets
longer, despite the fact that missing this drift has minimal
impact on training loss during the first 200 timesteps.

A reasonable concern would be that we are making this statement from
the advantageous position of studying a
well-understood phenomenon, and in a more open-ended task we would not
have the luxury of having prior knowledge of the correct
solution. However, we should note that the analysis is only partially
based on prior knowledge: The approximations that led to identifying
$\left(\sin(\theta_t)\cdot\omega_t/\Delta\omega_t\right)^2$
were guided by the network's structure and aimed at
establishing the validity of the result. In many open-ended tasks the
operator is very likely to have similar analytical or experimental
tools to test the validity of the dependencies between variables
hypothesised by the network.

Which brings us to the second major advantage of the KAN framework:
the \emph{controllability} afforded
to the operator to manually prune the dependencies that appear to be
coincidental. As explained in Section~\ref{sec:kan}, the training
algorithm is biased towards sparser, less-connected networks and also
provides indications about which connections can be safely pruned with
minimal loss in performance; This metric is depicted in
Figure~\ref{fig:seqkan_best_trained} as darker shades for the
connections that the network considers important. Now that we have
analysed the network and have established which connections are
qualitatively meaningful, we can easily prune those that are not,
especially those in lighter shades.\footnote{Note that the
  quantitative results in Table~\ref{tab:combined_inter_extrapolation}
  are before any such manual intervention.}

\begin{figure}[t]
\begin{center}
\begin{tikzpicture}[
    every node/.style={font=\tiny},
    block/.style={rectangle, draw, fill=blue!20, rounded corners, minimum width=1.6cm, minimum height=0.8cm},
    circleblock/.style={circle, draw, fill=black, minimum size=0.01cm},
    input/.style={coordinate},
    output/.style={coordinate},
    arrow/.style={->, thick},
    line/.style={-, thick},
    kaninput/.style={circle, draw, fill=red!30, minimum size=0.1cm},
    hidden/.style={circle, draw, fill=olive!30, minimum size=0.1cm},
    kanoutput/.style={circle, draw, fill=red!30, minimum size=0.1cm},
]

\node[] (phi1) {\includegraphics[width=6cm]{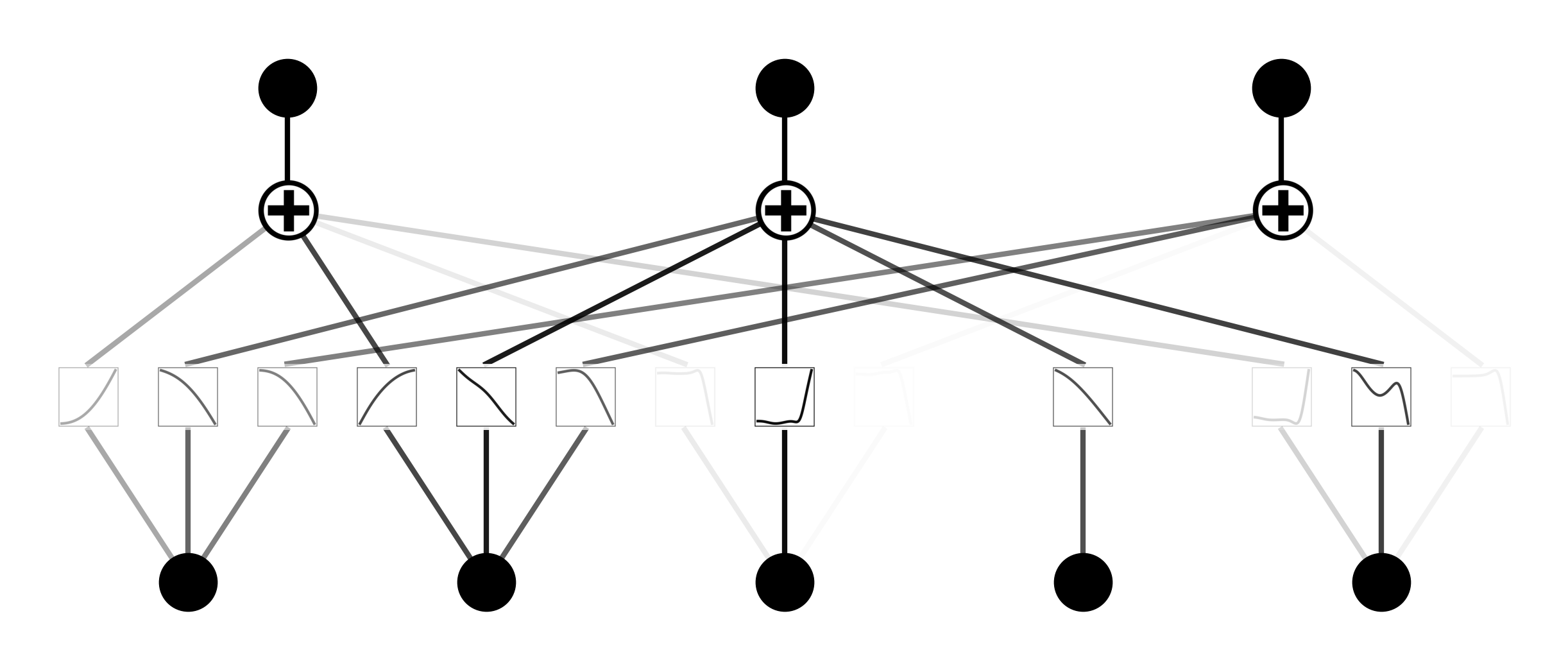}};

\node[above=-0.6cm of phi1] (phi2) [xshift=0cm, yshift=0.0cm] {\includegraphics[width=6cm]{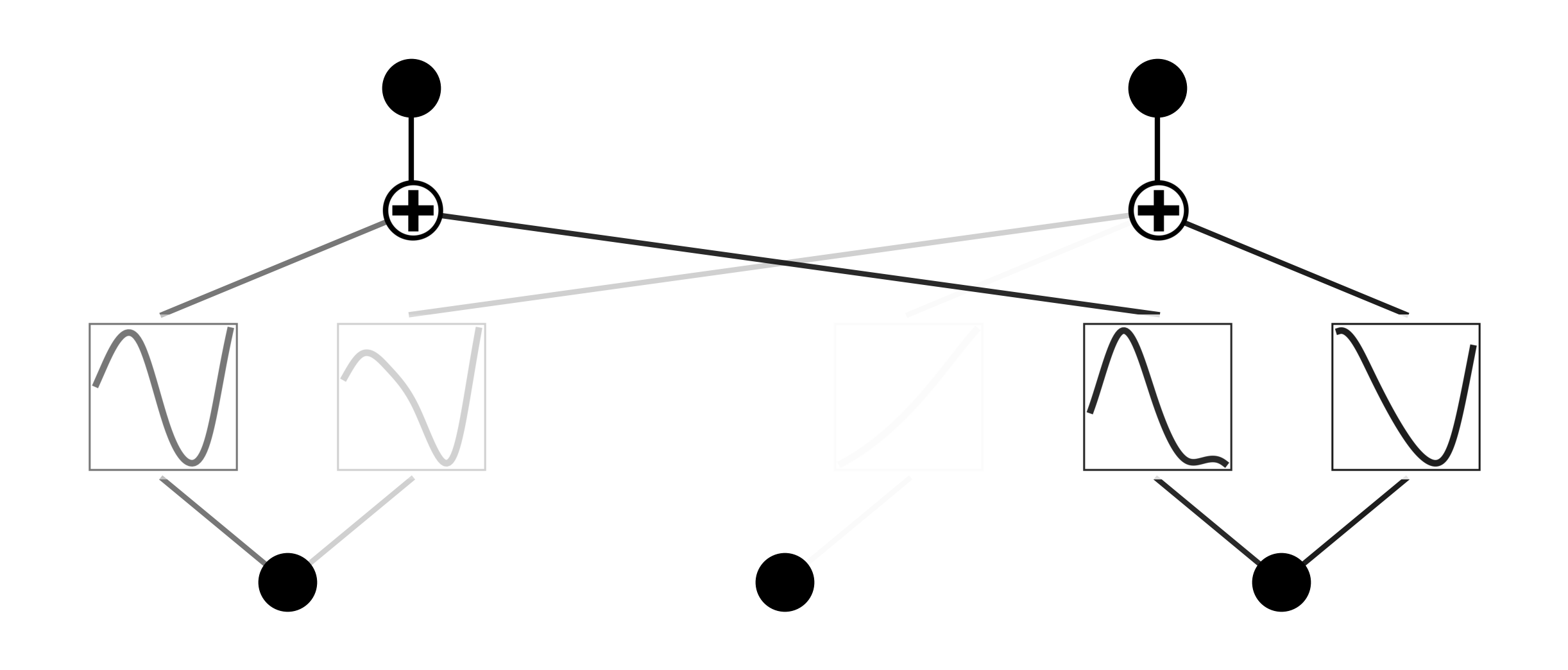}};

\node[kaninput, below=-0.6cm of phi1] (x0t) [xshift=-2.3cm, yshift=0.0cm] {$\theta_t$};
\node[kaninput, below=-0.6cm of phi1] (x1t) [xshift=-1.15cm, yshift=0.0cm] {$\omega_t$};
\node[hidden, below=-0.6cm of phi1, scale=0.65] (h0t1) [xshift=-0.0cm, yshift=0.0cm] {$h_0^{t-1}$};
\node[hidden, below=-0.6cm of phi1, scale=0.65] (h1t1) [xshift=1.75cm, yshift=0.0cm] {$h_1^{t-1}$};
\node[hidden, below=-0.6cm of phi1, scale=0.65] (h1t1) [xshift=3.5cm, yshift=0.0cm] {$h_2^{t-1}$};

\node[hidden, above=-0.6cm of phi1, scale=0.85] (h0t) [xshift=-2.25cm, yshift=0.0cm] {$h_0^{t}$};
\node[hidden, above=-0.6cm of phi1, scale=0.85] (h1t) [xshift=0.0cm, yshift=0.0cm] {$h_1^{t}$};
\node[hidden, above=-0.6cm of phi1, scale=0.85] (h1t) [xshift=2.25cm, yshift=0.0cm] {$h_2^{t}$};

\node[kaninput, above=-0.6cm of phi2, scale=0.75] (y0t) [xshift=-1.9cm, yshift=0.0cm] {$\Delta E_t$};
\node[kaninput, above=-0.6cm of phi2, scale=0.75] (y1t) [xshift=1.9cm, yshift=0.0cm] {$EQ_t$};

\node[draw, dashed, rounded corners, fill=blue!10, fit=(phi1), xscale=0.9, yscale=0.45, fill opacity=0.3, draw opacity=0.6] {};
\node[draw, dashed, rounded corners, fill=blue!10, fit=(phi2), xscale=0.85, yscale=0.45, fill opacity=0.3, draw opacity=0.6] {};
\end{tikzpicture}

\caption{Trained seqKAN/wide model. Unlike seqKAN, it takes only the
  current $\omega$ and $\theta$ as inputs and has three hidden-state
  nodes, giving it room to maintain the previous $\omega$ and $\theta$
  values if the data indicates that this is meaningful.}
\label{fig:seqkan_wide_trained}
\end{center}
\vspace{1.5em}
\end{figure}
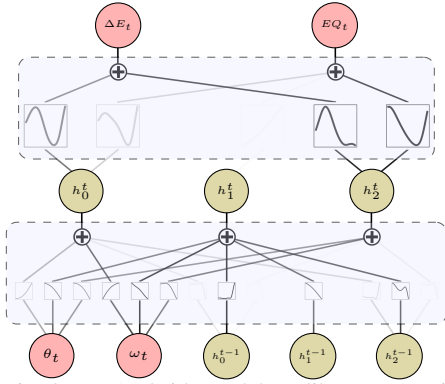

We can now direct our attention to the network learned by seqKAN/wide.
If we read Figure~\ref{fig:seqkan_wide_trained} we observe that
the Energy-Increasing output sums a function over $\omega_t$ with a
function over the sum $\omega_t+\theta_t$.
The network does not utilize any information
from the past as the hidden states do not contribute to the recurrency
and predictions are made solely based on the current timestep.
This analysis is corroborated by the quantitative results in
Table~\ref{tab:combined_inter_extrapolation}, where
we see that seqKAN/wide perfectly predicts the second label for which
only the current input is relevant.

A further observation is that seqKAN/wide `wasted' the extra degrees
of freedom afforded to its hidden-state layer to learn a different
function for each label instead of forcing the layers to learn
functions that contribute to both, as in seqKAN.  As a result, it
exploited the additional capacity to predict each label independently
leading to a distribution of the training loss to different functions,
rather than learning more intricate functions that are used for both
labels. In essence, the network used the extra capacity not to
discover that the inputs should also be carried verbatim to the output
layer besides contributing to the hidden-state layer, but to settle
for two unconnected theories, one for each label.

As final remark, we will comment on the fact that the additive
functions extracted from KAN networks of splines are a substantial
restriction on the shape of the theories that KANs can formulate,  and
the functions read out of a network need to be understood under this
perspective. More concretely, the additive expressions learned by
either \mbox{seqKAN} or seqKAN/wide are not, at the surface, similar
to the multiplicative expression
$\left(\sin(\theta_t)\cdot\omega_t/\Delta\omega_t\right)^2$ and
the quantitative performance of the seqKAN expression could be
considered coincidental. But we should read the comparative analysis
of the two networks as revealing that $\omega^2$ captures an essential
feature of the phenomenon, which is indeed true as the periods at
which squared angular velocity exhibits local maxima correlate with
the length of the pendulum string; By contrast, the
$\omega_t+\theta_t$ expression learned by seqKAN/wide does not.
We will revisit this point in our discussion of the capabilities and
limitations of KANs in Section~\ref{sec:conc}.

\subsection{Comparative Results}
\label{sec:exp:comp}

\begin{table*}[t]
    \centering
    \caption{Evaluation of the performance of RNN, LSTM, TKAN, seqKAN/wide and seqKAN models for the dual classification
    task with the pendulum length changing exponentially and sequence length being 10.
    The measured metrics are the areas under: (a) the Receiver Operating Characteristic curve (ROC-AUC)
    and (b) the Precision Recall curve (PR-AUC).}
    \vspace{1.5em}
    \label{tab:combined_inter_extrapolation}
    
    \begin{tiny}
        \begin{sc}

    \begin{subtable}[t]{0.48\textwidth}
        \centering
        \caption{Interpolation}
        \label{tab:baseline_interpolation}

        \begin{tabular}{l
            >{\columncolor{group1}}c >{\columncolor{group1_alt}}c
            >{\columncolor{group1}}c
            >{\columncolor{group2}}c >{\columncolor{group2_alt}}c}
            \toprule
            \multirow{2}{*}{\textbf{Metric}} & \multicolumn{5}{c}{\textbf{Architecture}} \\
            \cmidrule(lr){2-6}
            \textbf{(AUC)} & \textbf{RNN} & \textbf{LSTM} & \textbf{TKAN} & \textbf{seqKAN/wide} & \textbf{seqKAN} \\
            \midrule
            &  \multicolumn{5}{c}{\textbf{Energy Increasing Label}} \\
            \midrule
            \textbf{ROC} & 0.92 & 0.98 & 0.77 & 0.88 & 0.96  \\
            \textbf{PR}  & 0.80 & 0.97 & 0.83 & 0.84 & 0.95 \\
            \midrule
            & \multicolumn{5}{c}{\textbf{Close to Equilibrium Label}} \\
            \midrule
            \textbf{ROC} & 1.00 & 1.00 & 0.97 & 0.89 & 0.98 \\
            \textbf{PR}  & 1.00 & 1.00 & 0.93 & 0.66 & 0.97 \\
            \bottomrule
        \end{tabular}
    \end{subtable}
    \hfill
    \begin{subtable}[t]{0.48\textwidth}
        \centering
        \caption{Extrapolation}
        \label{tab:baseline_extrapolation}
        \begin{tabular}{l
            >{\columncolor{group1}}c >{\columncolor{group1_alt}}c
            >{\columncolor{group1}}c
            >{\columncolor{group2}}c >{\columncolor{group2_alt}}c}
            \toprule
            \multirow{2}{*}{\textbf{Metric}} & \multicolumn{5}{c}{\textbf{Architecture}} \\
            \cmidrule(lr){2-6}
            \textbf{(AUC)} & \textbf{RNN} & \textbf{LSTM} & \textbf{TKAN} & \textbf{seqKAN/wide} & \textbf{seqKAN} \\
            \midrule
            & \multicolumn{5}{c}{\textbf{Energy Increasing Label}} \\
            \midrule
            \textbf{ROC} & 0.70 & 0.68 & 0.71 & 0.56 & 0.90  \\
            \textbf{PR}  & 0.51 & 0.59 & 0.82 & 0.53 & 0.90  \\
            \midrule
            & \multicolumn{5}{c}{\textbf{Close to Equilibrium Label}} \\
            \midrule
            \textbf{ROC} & 0.72 & 1.00 & 0.79 & 1.00 & 0.98 \\
            \textbf{PR}  & 0.89 & 1.00 & 0.88 & 1.00 & 0.99 \\
            \bottomrule
        \end{tabular}
    \end{subtable}
             \end{sc}
        \end{tiny}
    \vspace{1.5em}
\end{table*}

In the previous section, we utilized the formulas that compute the labels of the predictive tasks to perform
an analysis on the relevance of the network's findings.
However, in machine learning terms, we have a dual classification task with two binary labels: whether the pendulum's
energy increases or whether it is close to the equilibrium point.
Thus, to compare the results across both interpolation and extrapolation tasks for different architectures we use
\term{Receiver Operating Characteristic-Area Under the Curve (ROC-AUC)} and
\term{Precision-Recall-Area Under the Curve (PR-CURVE AUC)} as the
evaluation metrics, since they provide a global assessment of the
models' performance without depending on specific thresholds. This choice aligns with
our goal of demonstrating the advantages of our model as a general
architecture, rather than a specific system solving a specific task.
However, when comparing our model to symbolic regression, we are
forced to use F1-score since symbolic regression does not output a
probability distribution.

In comparing the performance of various sequence processing algorithms, as shown in Table~\ref{tab:baseline_interpolation},
all models generalize well on the interpolation task. This is a positive and expected outcome, as it suggests that all
models are capable of capturing the core structure of the problem.
However, with the exception of seqKAN, all models perform significantly worse on the extrapolation task, particularly
when predicting the challenging Energy-Increasing label, as seen in Table~\ref{tab:baseline_extrapolation}. seqKAN, on the
other hand, maintains consistently high performance metrics, indicating that it is the only model that effectively recovers the hidden variable (string length).

The standard sequence processing algorithms, RNN and LSTM, perform similarly, with the LSTM exhibiting slightly better
performance. This is expected, as the LSTM, with its cell state, decides which information to retain from the recent or
distant past, whereas the RNN updates its history at every
timestep. Specifically in our task, we know for instance that
$\Delta\omega$ (two steps) is needed in order to make accurate
predictions. The LSTM is more capable of discovering such dependencies.

Regarding the TKAN architecture, which combines LSTM and KAN, it is interesting to note that it performs worse than the
vanilla LSTM on the interpolation task but similarly on the extrapolation task. Our intuition suggests that this occurs
because the addition of KANs to the LSTM’s output prevents the network
from overfitting, encouraging it to learn more meaningful
patterns. This could explain its consistently average performance on
both tasks, by comparison to LSTM's performance on the interpolated
test set caving in on the extrapolated test set.  However, we are
unable to test this hypothesis, as TKAN is based on an implementation
of the KAN layer that does not make the learned functions available
for plotting.\footnote{This is noted as a known issue in the TKAN
  repository that was unresolved as of writing this text.}

In our selected task, while symbolic regression is not expected to perform perfectly due to the lack of an analytic
solution, it is anticipated to provide relevant results based on its prior knowledge of potential functions for
approximating the phenomenon. However, as shown in Table~\ref{tab:symb_reg_inter_extrapolation}, symbolic regression
struggles with the difficult task of predicting the increasing energy change in both interpolation and extrapolation
scenarios. The symbolic method produces highly intricate formulas that
fit the training data too tightly to generalize well. Additionally, in
each run, the functions change significantly, which is expected due to
the absence
of a closed form. These results further support the idea that seqKAN is particularly well-suited for modeling temporal
formulas with a mathematical structure, even in the absence of an analytical solution.

\begin{table}[t]
    \centering
    \caption{Evaluation of Symbolic Regression (SR) and seqKAN. In SR, we approximate the functions of
    the energy, the $\omega_t$, and the $\theta_t$ and utilize them to derive the final labels. We present the best
    performing SR model.
    The evaluation metric used is
    the F1 score. In the case of seqKAN, the optimal F1 score was computed from the thresholds of the
    PR curve. For the first label, the
    percentage of seqKAN's PR curve where its F1-score exceeds that of SR is  98.86\% in both
    interpolation and extrapolation cases.}
    \label{tab:symb_reg_inter_extrapolation}
 
    \vspace{1.5em}

    \begin{tiny}
        \begin{sc}

    \begin{subtable}[t]{0.48\textwidth}
        \centering
        \caption{Interpolation}
        \label{tab:symb_reg_interpolation}
        \begin{tabular}{lcc}
            \toprule
            \multirow{2}{*}{\textbf{Metric}} & \multicolumn{2}{c}{\textbf{Architecture}}\\
            \cmidrule(lr){2-3}
            & \textbf{Symbolic Regression} & \textbf{seqKAN} \\
            \midrule
            \midrule
            \multirow{4}{*}{\textbf{F1 Score}} & \multicolumn{2}{c}{\textbf{Energy Increasing Label}} \\
            \cmidrule(lr){2-3}
             & 0.47 &  0.95\\
            \cmidrule(lr){2-3}
            & \multicolumn{2}{c}{\textbf{Close to Equilibrium Label}} \\
            \cmidrule(lr){2-3}
             & 1.0  & 0.92  \\
            \bottomrule
        \end{tabular}
    \end{subtable}
    \vfill
    \vspace{1.5em}
    \begin{subtable}[t]{0.48\textwidth}
        \centering
        \vspace{1em}
        \caption{Extrapolation}
        \label{tab:symb_reg_extrapolation}
        \begin{tabular}{lcc}
            \toprule
            \multirow{2}{*}{\textbf{Metric}} & \multicolumn{2}{c}{\textbf{Architecture}}\\
            \cmidrule(lr){2-3}
            & \textbf{Symbolic Regression} & \textbf{seqKAN} \\
            \midrule
            \midrule
            \multirow{4}{*}{\textbf{F1 Score}} & \multicolumn{2}{c}{\textbf{Energy Increasing Label}} \\
            \cmidrule(lr){2-3}
             & 0.48 &  0.89\\
            \cmidrule(lr){2-3}
            & \multicolumn{2}{c}{\textbf{Close to Equilibrium Label}} \\
            \cmidrule(lr){2-3}
             & 1.0  & 0.95  \\
            \bottomrule
        \end{tabular}
    \end{subtable}
             \end{sc}
        \end{tiny}
    \vspace{1.5em}
\end{table}

\section{Conclusions and Future Work}
\label{sec:conc}

In this paper we presented \term{seqKAN}, a recurrent KAN architecture
for processing sequences. seqKAN is more faithful to the core concept
of the KAN framework than alternative proposals for sequence
processing as it ports \term{concepts} from the RNN framework to KAN
without directly integrating processing cells or other MLP-inspired
structures.

This approach is motivated by the need to retain the interpretability
and controllability offered by the less distributed nature of KAN
parameters, which are diluted when part of the knowledge distilled
from the data is represented in highly-distributed MLP
representations. In order to demonstrate this advantage, we
qualitatively analysed two alternative seqKAN architectures applied on
the same sequence processing task, and offered a human-understandable
explanation of dependencies between the variables of the task each of
the two networks learned.

What is important is that this interpretability was \emph{not}
achieved at the expense of quantitative performance, but combined with
performance superior to TKAN (an alternative architecture for
KAN-based sequence processing), RNN, LSTM, and symbolic
regression. The key characteristic of the evaluation task was that it
was not stationary but was influenced by an underlying phenomenon
represented by a variable that was not directly visible but that could
be inferred from the input. From this phenomenon we extracted an
interpolation test-set where the hidden variable was within the same
value range as during training and an extrapolation test-set where the
hidden variable was outside the range seen during training (although
still observing the same natural laws). With this experimental setup,
we demonstrated that the performance gap between our approach and the
systems we compared against \emph{increased} on the extrapolation
test-set.

Having said that, the KAN framework (and consequently seqKAN) poses
limitations on the theories that can be represented. The most obvious
such limitation, as discussed in the final remark of
Section~\ref{sec:exp:seqkan}, is the additive form of the network.
Although KANs are inspired by the universality of the
Kolmogorov-Arnold Theorem, for the reasons discussed in
Section~\ref{sec:kan} the representation based on addition and splines
is in reality an approximation. \citet[Section~2]{liu-wang-etal:2024}
show that one can achieve arbitrarily low error by stacking multiple
KAN layers, but such an approach weakens the clarity of the result and
reduces the confidence that a real, causal dependency has been
discovered.

This limitation shows us three directions of possible future research:
mapping its extent and impact, maximizing the value we can extract
despite it, and lifting it. Regarding the first, the KAN framework
has attracted considerable attention in a very short time,
so we are confident that a community will coalesce around applying
KANs to different tasks so that empirical experience can accumulate.
Our immediate plan in this respect is to also develop a
Transformer-inspired KAN sequence processor and compare seqKAN and
the new architecture on both generated and real-world tasks.
What is important for such experiments is reading the expressions
learned by the networks and reporting on the impact of this
representational limitation.

Which relates to the second future research path, which is to develop
human-computer interaction methods that can assist with the
painstaking process of extracting symbolic expressions from KAN
networks. As there is a delicate balance between over-simplifying what
is represented in the network and overwhelming the operator with
details, this should be approached in a systematic way involving
test operators with varying expertise on KANs.

The final, and most ambitious, goal of generalizing the framework to
allow for more complex expressions runs the risk of pushing the
framework towards the direction of symbolic regression. Symbolic
regression has its own goals and application areas, which are
distinct from those of KANs; KAN aims to be a \term{connectionist}
and not a symbolic machine learning framework, and its main
juxtaposition is the extremely distributed representation of MLP.
In other words, to admin more complex expressions the framework needs
to first carry out research on how to prioritize simple addition of
learned curves admitting expression complexity only as a last resort.
This is an ambitious goal which might require re-thinking foundational
aspects of the framework such how back-propagation distributes
loss.

\appendix
\section{Pendulum with varying string length}
\label{sec:pendulum}

\begin{figure}[tb]
    \begin{center}
        \begin{tikzpicture}
            \pgfmathsetmacro{\myAngle}{30}

            \coordinate (centro) at (0,0);
            \draw[dashed,gray,-] (centro) -- ++ (0,-3.25) node (mary) [midway,left]{};
            \draw[thick] (centro) -- ++(270+\myAngle:3) coordinate (bob) node[near end,midway, right]{$L$};
            \draw[gray] (centro) -- ++(-150+\myAngle:2.5) coordinate (bob1) node[near end,midway, right]{};
            \pic [draw, ->, "$\theta$", angle eccentricity=1.5] {angle = mary--centro--bob};
            \draw [-stealth] (bob) -- ++(0,-1.0)
              coordinate (g)
              node[near end,left] {$g$};

            \draw[->, thick] (bob) arc[start angle=180, end angle=-120, radius=0.2] node[midway, right] {$\omega$};
            \draw[dotted, thick, gray] (bob1) arc[start angle=\myAngle, end angle=+(180-\myAngle), radius=-1.45];
            %\draw[dotted, thick, gray] (bob) arc[start angle=-\myAngle, end angle=-(180-\myAngle), radius=1.7];
            \draw[dotted, thick, gray] (bob) arc[start angle=-(\myAngle+15), end angle=-(180-\myAngle), radius=1.7];

            \filldraw [fill=black!40,draw=black] (bob) circle[radius=0.1];
            \filldraw [fill=black!20,draw=black!40] (bob1) circle[radius=0.1];
        \end{tikzpicture}
        \caption{The illustration shows a pendulum with variable length. The equilibrium point is where the string is vertical.
           $\theta$ represents the displacement from equilibrium, while $\omega$ is the angular velocity. The pendulum swings
            between two extreme positions, showing its range of motion, and the length  $L$ of the string changes as the
            pendulum moves, making it time dependent as well.}
        \label{fig:pendulum}
    \end{center}
\vspace{1.5em}
\end{figure}
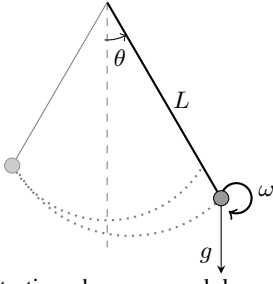

A \emph{pendulum} is a mass (referred to as `bob') attached to the end
of a string that swings back and forth in a periodic motion under the
influence of gravity. The motion of the pendulum is characterized by
two key components: the angular displacement $\theta$ of the pendulum
from the vertical and the angular velocity $\omega$, which is the
first derivative of $\theta$ (Figure~\ref{fig:pendulum}).

The motion of a pendulum can be described using a set of equations
derived from Newton's second law of motion and the principles of
rotational dynamics. To model the pendulum's motion over time, we
first compute the \emph{angular acceleration} \(\alpha_t\) which is
proportional to the restoring torque due to gravity:
$$\alpha_t = -\frac{g}{L_t} \sin(\theta_t)$$
where \(g\) is the acceleration due to gravity, and \(L_t\) is the
length of the pendulum at that specific time.

The \emph{angular velocity} \(\omega_t\) and the angular
displacement \(\theta_t\) are then incrementally computed from their
initial values using numerical integration:
\[
\begin{array}{c}
\omega_t = \omega_{t-1} + \alpha_t \cdot \Delta t \\
\theta_t = \theta_{t-1} + \omega_t \cdot \Delta t \\
\end{array}
\]
If we substitute for $a_t$ in the first equation, we can also see that
string length can be calculated from displacement and the difference
in velocities between consecutive time steps:
\[
\begin{array}{c}
\displaystyle
\Delta\omega = -\frac{g}{L_t} \sin(\theta_t) \cdot \Delta t \\
\displaystyle
L_t = -\frac{g\sin(\theta_t)}{\Delta\omega / \Delta t} \\
\end{array}
\]
%
%Since KANs favour additive representations, we expect a KAN to
%calculate $\log{L}$ instead of $L$:
%
%\[
%\log{L_t} = -\mathrm{sign}(\theta_t)\log\left|g\sin(\theta_t)\right| 
%            +\mathrm{sign}(\Delta\omega)\log\frac{\left|\Delta\omega\right|}{\Delta t} \\
%\]
%

Our Close-to-Equilibrium binary label is defined be the condition:
$$ \theta \cdot \omega \geq 0$$
The total mechanical energy of the pendulum is the sum of kinetic and
potential energies, if we assume the mass is concentrated at the end
of the pendulum:
\begin{align*}
E_t & = & \frac{1}{2} & m\left(L_t\omega_t\right)^2 & - \quad & m g L_t \cos(\theta_t) \\
    & = &
\frac{1}{2} &m \left(\frac{g\sin(\theta_t)\omega_t}{\Delta\omega / \Delta t}\right)^2
    & + \quad &
\frac{mg^2 \sin(\theta_t)\cos(\theta_t)}{\Delta\omega / \Delta t} \\
    & = &
\frac{1}{2} &m \left(\frac{g\sin(\theta_t)\omega_t}{\Delta\omega / \Delta t}\right)^2
    & + \quad &
\frac{mg^2 \sin(2\theta_t)}{2\Delta\omega / \Delta t} \\
\end{align*}
Our Energy-Increasing binary label can then be computed as follows,
assuming unit step for all $\Delta t$ since all our time steps are equal:
\[
\begin{array}{c}
\displaystyle
\Delta\left(\frac{\sin(\theta_t)\omega_t}{\Delta\omega_t}\right)^2
 + \Delta\left(\frac{\sin(2\theta_t)}{\Delta\omega_t}\right)
 \geq 0 \\
\end{array}
\]

\begin{figure}[t]
\begin{center}
\includegraphics[width=\linewidth]{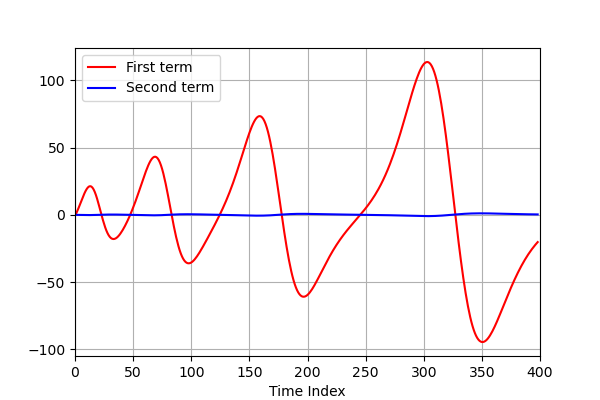}
\caption{Plot of the two terms in the expression of the
  Energy-Increasing label. As can be clearly seen, the sum of the two
  terms is completely dominated by the first term.}
\label{fig:energy_diff}
\end{center}
\vspace{1.5em}
\end{figure}

As can be seen in Figure~\ref{fig:energy_diff}, the sign of the sum of
these two $\Delta$ terms is completely dominated by the first term.
This indicates that a data-driven (as opposed to analytical) model
for the Energy-Increasing label that completely ignores the second
term should be considered correct.

\end{document}